\documentclass[letterpaper, 10 pt, conference]{format/ieeeconf}

\usepackage{amsmath}
\usepackage{amssymb}
\usepackage{amsfonts}

\usepackage{amsthm}
\usepackage[ruled,vlined]{algorithm2e}
\usepackage{mathtools}
\usepackage{tabularx}
\usepackage{graphicx}
\usepackage{subcaption} 
\usepackage{enumerate}
\usepackage{booktabs}
\usepackage{float}
\usepackage{url}
\usepackage{verbatim}
\usepackage{booktabs} 
\usepackage{multirow}
\usepackage{color}
\usepackage{pifont}
\usepackage[linkcolor=black,citecolor=black,urlcolor=black,colorlinks=true]{hyperref}
\usepackage{cite}
\usepackage{mathrsfs}

\graphicspath{{figures/}}
\IEEEoverridecommandlockouts
\title{An Efficient Trajectory Planner for\\ Car-like Robots on Uneven Terrain}

\author{Long Xu$^{1,2}$, Kaixin Chai$^{2,3}$, Zhichao Han$^{1,2}$, Hong Liu$^{4}$,  \\Chao Xu$^{1,2}$, Yanjun Cao$^{2}$, and Fei Gao$^{1,2}$
\thanks{This work was supported by the Fundamental Research Funds for the Central Universities and the National Natural Science Foundation of China under grant no. 62003299.}
\thanks{$^{1}$State Key Laboratory of Industrial Control Technology, Zhejiang University, Hangzhou 310027, China.}
\thanks{$^{2}$Huzhou Institute of Zhejiang University, Huzhou 313000, China.}
\thanks{$^{3}$Institute of Artificial Intelligence and Robotics, Xi'an Jiaotong University, Xi'an 710049, Shaanxi, China.}
 \thanks{$^{4}$School of information and electrical engineering, Hangzhou City University, Hangzhou 310015, China. }
	\thanks{\textit{Corresponding authors: Hong Liu, Fei Gao}. E-mail: {\tt\small \{gaolon, fgaoaa\}@zju.edu.cn, liuhong@zucc.edu.cn}}
}

\begin{document}

    \maketitle
    \thispagestyle{empty}
    \pagestyle{empty}

\begin{abstract}
Autonomous navigation of ground robots on uneven terrain is being considered in more and more tasks. However, uneven terrain will bring two problems to motion planning: how to assess the traversability of the terrain and how to cope with the dynamics model of the robot associated with the terrain. The trajectories generated by existing methods are often too conservative or cannot be tracked well by the controller since the second problem is not well solved. In this paper, we propose \emph{terrain pose mapping} to describe the impact of terrain on the robot. With this mapping, we can obtain the $SE(3)$ state of the robot on uneven terrain for a given state in $SE(2)$. Then, based on it, we present a trajectory optimization framework for car-like robots on uneven terrain that can consider both of the above problems. The trajectories generated by our method conform to the dynamics model of the system without being overly conservative and yet able to be tracked well by the controller. We perform simulations and real-world experiments to validate the efficiency and trajectory quality of our algorithm.
\end{abstract}

\section{Introduction}
\label{sec:Introduction}

An increasing number of tasks require ground robots to navigate autonomously on uneven terrain, such as forest rescue, wilderness exploration, mining transportation, etc. As well as localization, mapping, and control, motion planning is a crucial part of autonomous navigation systems. Existing 2D indoor navigation techniques for ground robots are relatively mature, and there are many open-source, practical motion planning algorithms\cite{teb1,teb2,ompl1}. However, none of them can be directly adapted to uneven terrain, since most of them ignore the essential fact that some properties of the terrain (such as height and curvature) will change with spatial location. Considering this fact mainly brings two new problems for motion planning:
\begin{itemize}
    \item[1)]How do we measure and consider the traversability of terrain in motion planning? 
\end{itemize}

For example, in general, we want the robot to travel on flat terrain instead of steep areas; even for similarly traversable terrain, one with firm, gentle soil while the other filled with rough gravel, we prefer the robot to travel on the former since the latter is more rugged and driving on it may be detrimental to other modules of the robot. Thus, the slope and roughness of the terrain should be considered in the planner.
\begin{itemize}
    \item [2)] How should we deal with the dynamics model of the robot associated with the terrain?
\end{itemize}

\begin{figure}	
    \centering
    \includegraphics[width=1.0\columnwidth]{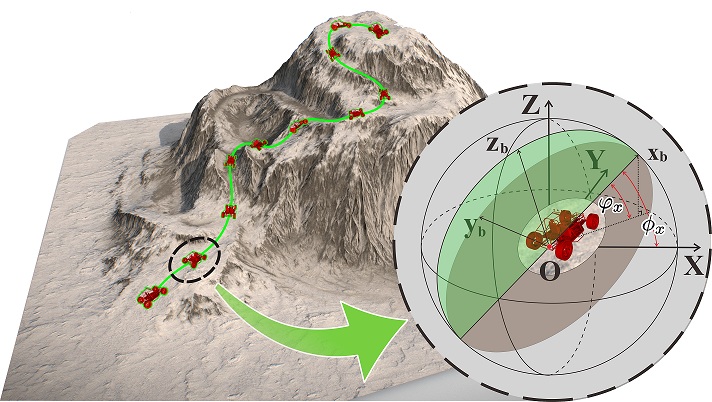}
    \caption{A car-like robot driving on uneven terrain, the green line indicates the trajectory optimized by the proposed method.}
    \label{fig:head}	
    \vspace{-0.3cm}
\end{figure}

As shown in the bottom right corner of Fig. \ref{fig:head}, suppose the car-like robot is driving on a sloping surface. The throttle required for the robot to reach the same acceleration in the body frame is different for uphill and downhill due to the presence of gravity. If the planner ignores the terrain in the dynamics model, the planned trajectory may be infeasible for the robot. Thus, it is necessary to handle the dynamics coupled with the terrain so that the controller can better track the trajectory.

An effective planner for robots on uneven terrain should consider both of these problems. The first problem is related to the safety of the robot, and the second problem is related to the executability of the trajectory. Attributed to the nonlinearity of the terrain, the rising dimensionality of the problem due to the high-dimensional robot state space, and the coupling of the robot dynamics model with the changing terrain, most existing work\cite{drivingonpoint,fastlocal,putn,travelhybrid,fastmesh,towards} cannot address the second problem well. As a result, the trajectories generated by these methods are often too conservative or cannot be tracked well by the controller.

To compensate for the shortcomings of existing methods due to the above factors, in this paper, through a mapping for describing the interaction between the terrain and the robot, we design an efficient optimization-based planning framework for car-like robots, which allows to considering both the first and the second problem effectively. The trajectories generated by our method conform to the dynamics model of the system without being overly conservative and yet able to be tracked well by the controller.

We model the motion planning problem of car-like robots on uneven terrain as an optimal control problem. Then, we propose \emph{terrain pose mapping} to describe the impact of terrain on the robot. With this mapping, we can obtain the $SE(3)$ state of the robot on uneven terrain for a given state in $SE(2)$. Also, we will show an algorithm that approximately constructs this mapping. Finally, we use piecewise polynomials to represent the trajectory and simplify the problem to a nonlinear constrained optimization problem containing constraints of non-holonomic dynamics, curvature, etc, which can be solved using numerical optimization algorithms. Besides, We perform comprehensive tests in simulation and real world to validate our method. Contributions of this paper are:

\begin{itemize}
\item [1)]We propose \emph{terrain pose mapping} to describe the impact of terrain on the robot and present an efficient algorithm to construct it approximately.
\item [2)]We propose an optimization-based planning framework for car-like robots on uneven terrain, which allows considering terrain curvature and dynamics of the robot.
\item [3)]We open source our software\footnote{\url{https://github.com/ZJU-FAST-Lab/uneven_planner}} for the reference of the community.
\end{itemize}

\section{Related Works}
\label{sec:RelatedWork}

\subsection{Motion Planning for Car-like Robots}
\label{subsec:Motion Planning for Car-like Robots}
The popularity of autonomous driving has extensively promoted the research of motion planning for car-like robots. Existing approaches can be roughly divided into sampling-based and optimization-based methods. The former is represented by RRT*\cite{rrtstar} and its variants, which ensure global optimality with sufficiently dense sampling. For the non-holonomic constraint of car-like robots, many algorithms that consider the nonlinear dynamics of the robot while sampling have been proposed\cite{kinorrt1,kinorrt2,kinorrt3,kinorrt4}, enhancing the efficiency of sampling-based planners in more scenarios. Although sampling-based methods can avoid local minima in non-convex environments, they confront a dilemma between computation overhead and trajectory quality.

Optimization-based approaches\cite{opt1,opt2,opt3,opt4,opt5,opt6} usually model the motion planning problem of a car-like robot as an optimal control problem (OCP) and represent the trajectory with discrete state points, then simplify the problem to nonlinear model predictive control (NMPC) problem\cite{opt4} or simpler quadratically constrained quadratic programming, quadratic programming\cite{opt1,opt2}, which can be solved by numerical optimization algorithms.

\subsection{Motion Planning on Uneven Terrain}
\label{subsec:Autonomous Navigation on Uneven Terrain}
In recent years, many works are trying to solve the two problems mentioned in Sec.\ref{sec:Introduction}, and most of them focus on the first one by combining various geometric information\cite{drivingonpoint,fastlocal,putn,travelhybrid,fastmesh}, exploiting conditional value-at-risk\cite{step}, fusing semantic information\cite{tns,jphow}, etc. For the second problem, most of them either avoid it in planning\cite{fastmesh,fastlocal} or consider only the geometric properties of the trajectory\cite{drivingonpoint,travelhybrid}, leaving the other problems to the controller.

Kr{\"u}si et al.\cite{drivingonpoint} presented a practical approach to global motion planning and terrain assessment for car-like robots in generic 3D environments, which assesses the traversability of terrain on demand during motion planning. However, in the face of complex environments, this work often requires a huge number of samples to generate trajectories, which does not guarantee real-time performance. To address the shortcomings of this method, Jian et al.\cite{putn} proposed a plane-fitting based uneven terrain navigation framework, which utilizes informed-RRT*\cite{rrtinformed} as the front end. It refines the path using Gaussian Process Regression\cite{gps} and finally generates dynamically feasible local trajectories by solving an NMPC problem. Although this work considers the changing terrain within NMPC, their method is inefficient in generating long trajectories because the complexity of the NMPC problem will rise dramatically as the planning horizon becomes longer. In order to assess the traversability more precisely, Zhang et al.\cite{suspension} considers the suspension system when the robot is stationary, but it requires an accurate identification of the robot's physical parameters and estimation of landing points of the four tires, which is less efficient.

The work \cite{elevation} uses surfel to represent the points cloud, incorporating both kinematic and physical constraints of robots, enabling efficient sampling-based planners for challenging navigation on uneven terrain. Nevertheless, the generated trajectory does not contain information about the robot's speed and acceleration with respect to time. Besides, when employing Dubins or Reeds Sheep state spaces, this method takes too long to converge to near-optimal paths. Wang et al.\cite{towards} proposed an optimization-based planning framework for ground robots considering both active and passive height changes on the z-axis. The trajectories planned by their method can provide dynamical information related to time and have excellent smoothness, benefiting from the penalty field for chassis motion constraints defined in $\mathbb{R}^3$. Although this method considers the velocity and acceleration of the trajectory in 3D space, the optimized trajectory is often dynamic infeasible or too conservative as the dynamics coupled with the changing terrain is not taken into account.

\section{Planning Framework}
\label{sec:Planning Framework}
A general motion planning problem for ground robots on uneven terrain can be expressed as the following optimal control problem:
\begin{align}
&\min_{\textbf{u}(t)}\int_0^{t_f}\tau(\textbf{s}(t),\textbf{s}^{(1)}(t),...,\textbf{s}^{(s)}(t),\textbf{u}(t))dt+\rho(t_f)\\
&s.t.\ \ \dot{\textbf{s}}(t)=f(\textbf{s}(t),\textbf{u}(t)),\label{con:statediff}\\
&\quad \quad \text{Ter}(\textbf{s}(t))=0,\label{con:tercontact}\\
&\quad \quad C_p(\textbf{s}(t),\textbf{s}^{(1)}(t),...,\textbf{s}^{(s)}(t),\textbf{u}(t))\preceq\textbf{0},\label{con:constraints}
\end{align}
where $\textbf{s}$ is the state of the robot,  $\textbf{s}^{(*)}$ is the $*$-th order derivative of  $\textbf{s}$, $\textbf{u}$ is the control input of the robot, $\rho:[0,\infty)\mapsto[0,\infty)$ is the time regularization. $\tau(*)$ denotes the cost associated with the task, such as energy, risk, etc. Eq.(\ref{con:statediff}) and Eq.(\ref{con:tercontact}) denote the state transfer equation and terrain contact constraint, respectively. In this work, we assume that the robot must be in contact with the terrain and its wheels do not slip. Eq.(\ref{con:constraints}), where $C_p(*):*\mapsto\mathbb{R}^{N_p}$ and $\textbf{0}=[0,0,..,0]^\text{T}\in\mathbb{R}^{N_p}$, denote $N_p$ inequality constraints of the robot, including dynamics constraints, traversability constraints, etc.

In this section, we will present how we simplify the state description of a car-like robot on uneven terrain by a mapping $\mathscr{F}$ called \emph{terrain pose mapping}, and give a specific algorithmic procedure for constructing this mapping. Then, we parameterize the state trajectory as piecewise polynomials and give explicit expressions for the control inputs, dynamical variables with the help of this mapping and the state transfer equation. Since $\tau(*)$ may differ from task to task, in this paper, we propose an exemplary cost function that combines terrain curvature and trajectory smoothness. Finally, we will analyze the more specific trajectory optimization problem after these processes, which can be solved using numerical optimization algorithms.

\subsection{Terrain Pose Mapping}
\label{subsec:Terrain Pose Mapping}

Using the simplified bicycle model\cite{bicycle} and describing the state of a car-like robot on uneven terrain by a $SE(3)$ state with the position $\textbf{p}=[x,y,z]^\text{T}\in \mathbb{R}^3$ and the attitude $\textbf{R}=[\textbf{x}_b,\textbf{y}_b,\textbf{z}_b]\in SO(3)$, we can write the robot's model as follows:
\begin{align}
&\dot{\textbf{p}}=\textbf{x}_b\cdot v_x,\label{model:pdot}\\
&\dot{\textbf{R}}=\textbf{R}\lfloor \frac{v_x\tan \delta}{L_w}\cdot\textbf{z}_b \rfloor\label{model:rdot},
\end{align}
where $v_x$ is the velocity along the body axis $\textbf{x}_b$, $\delta$ is the steering angle, $L_w$ is the wheelbase length of the robots, the operation $\lfloor*\rfloor$ takes a vector to a skew-symmetric matrix. It is worth noting that on flat ground, the state of the robot can be represented using only $\textbf{s}_r=[x,y,\theta]^\text{T}\in SE(2)$. However, due to the presence of uneven terrain, we need to use a higher dimensional representation for the state.

In this work, we consider the terrain contact constraint (\ref{con:constraints}) brings about actually a mapping relation called \emph{terrain pose mapping} $\mathscr{F}:SE(2)\mapsto\mathbb{R}\times\mathbb{S}^2_+$, where $\mathbb{S}^2_+\triangleq\{\textbf{x}\in\mathbb{R}^3\ |\ \lVert\textbf{x}\rVert_2=1,\textbf{x}\cdot\textbf{b}_3>0\},\textbf{b}_3=[0,0,1]^\text{T}$. This means that the state in $SE(2)$, which is originally on flat ground, is given height and attitude, with elements in $\mathbb{R}$ representing the height $z$, elements in $\mathbb{S}_+^2$ representing the body axis $\textbf{z}_b$. It is worth noting that we do not use the entire 2D sphere $\mathbb{S}^2\triangleq\{\textbf{x}\in\mathbb{R}^3\ |\ \lVert\textbf{x}\rVert_2=1\}$, but $\mathbb{S}_+^2$, because in a common task, we do not want the body axis $\textbf{z}_b$ to face the lower half-plane, which requires high speed and is not safe.

Let $\textbf{x}_{yaw}=[\cos\theta,\sin\theta,0]^\text{T}$ denote the direction of yaw angle, the mapping $\mathscr{F}$ can be expressed as two functions:
\begin{align}
&z=f_1(x,y,\theta),\\
&\textbf{z}_b=\textbf{f}_2(x,y,\theta)\label{con:zb}.
\end{align}
Using the Z-X-Y Euler angles to represent the attitude of the robot, we can obtain:
\begin{align}
&\textbf{p}=[x,y,z]^\text{T}=[x,y,f_1(x,y,\theta)]^\text{T},\\
&\textbf{y}_b=\frac{\textbf{f}_2(x,y,\theta)\times\textbf{x}_{yaw}}{\lVert\textbf{f}_2(x,y,\theta)\times\textbf{x}_{yaw}\rVert},\\
&\textbf{x}_b=\textbf{y}_b\times\textbf{f}_2(x,y,\theta).
\end{align}
Thus, the terrain contact constraint allows us to still use elements in $SE(2)$ to describe the state of the robot on uneven terrain.

Many methods can be used to construct the mapping $\mathscr{F}$; the most accurate method is to remotely drive the robot onto real terrain, collect accurate localization data and fit it, but this method is ineffective for regions that the robot has not reached. Usually, it is easier for the robot to obtain the point cloud of the environment by some LIDAR-based SLAM algorithms\cite{fast-lio2}. Thus, in this paper, we propose a simple and efficient method to obtain the mapping $\mathscr{F}$ from the point cloud by using an iterative plane-fitting strategy while considering the size and attitude of the robot. The pipeline for processing each $SE(2)$ state is shown in Algorithm \ref{alg:mapping construct}.

\begin{algorithm}
    \caption{Get the result of $\mathscr{F}$ at a $SE(2)$ state }
    \label{alg:mapping construct}
    \KwIn{state $\textbf{s}_r\in SE(2)$, Iteration times $N_{iter}$, Ellipsoidal parameters $(e_x,e_y,e_z)$}
    \KwOut{$\textbf{z}_b$,\ z}
    \Begin
    {
        $\textbf{z}_b\leftarrow\textbf{b}_3$\;
        $z\leftarrow\text{FindNearestXYPointZ}(\textbf{s}_r)$\;
        \For {\textbf{each} $i\in N_{iter}$ }
        {
            $(p_i,R_i)\leftarrow\text{CalculateSE3}(\textbf{s}_r,z,\textbf{z}_b)$\;
            $G_i\leftarrow\text{FindEllipsoidPoints}(p_i,R_i,e_x,e_y,e_z)$\;
            $p_{mean}\leftarrow\text{GetMeanPosition}(G_i)$\;
            $Cov\leftarrow\text{ZeroSquareMatrix3}()$\;
            \For{\textbf{each} $p_j \in G_i $}
            {
                $p_e\leftarrow p_j-p_{mean}$\;
                $Cov\leftarrow Cov+p_ep_e^\text{T}$\;
            }
            $p_{mean}\leftarrow p_{mean}/\text{NumOf}(M_i)$\;
            $Cov\leftarrow Cov/\text{NumOf}(M_i)$\;
            $\textbf{z}_b\leftarrow\text{GetMinEigenVec}(Cov)$\;
            $z\leftarrow p_{mean}.\text{GetZ}()$\;
        }
        \textbf{return} $\textbf{z}_b$,\ z\;
    }
\end{algorithm}

For a $SE(2)$ state, $\textbf{z}_b$ is first initialized to $\textbf{b}_3$, then we search the point cloud for a neighboring point whose $(x,y)$ coordinate is nearest, and use the height of this point as the initial value of $z$. Thus, we can obtain the corresponding $SE(3)$ state of the robot. Next, the points in an ellipsoidal region related to the robot's size, position, and attitude will be taken out. And so, $\textbf{z}_b$ will be updated by the eigenvector corresponding to the smallest eigenvalue of their covariance matrix, $z$ will be updated by the average of the heights of the points, which allows us to obtain the new ellipsoid region for the next iteration.

\begin{figure}	
    \centering
    \subcaptionbox{process of the construction.}{\includegraphics[width=1.0\columnwidth]{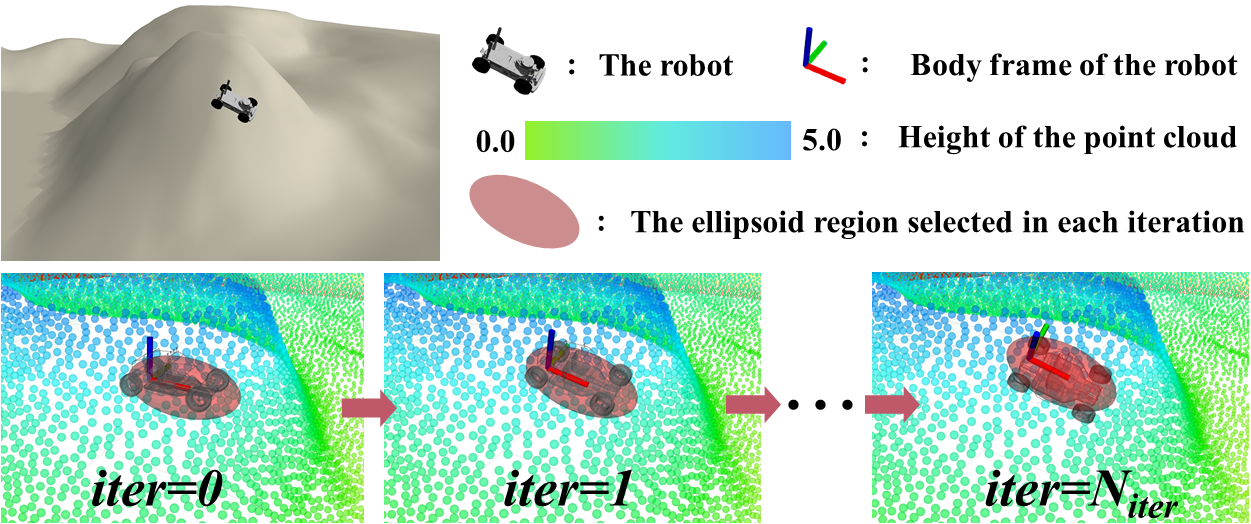}}
    \subcaptionbox{when $\theta=\theta_0$.}{\includegraphics[width=0.49\columnwidth]{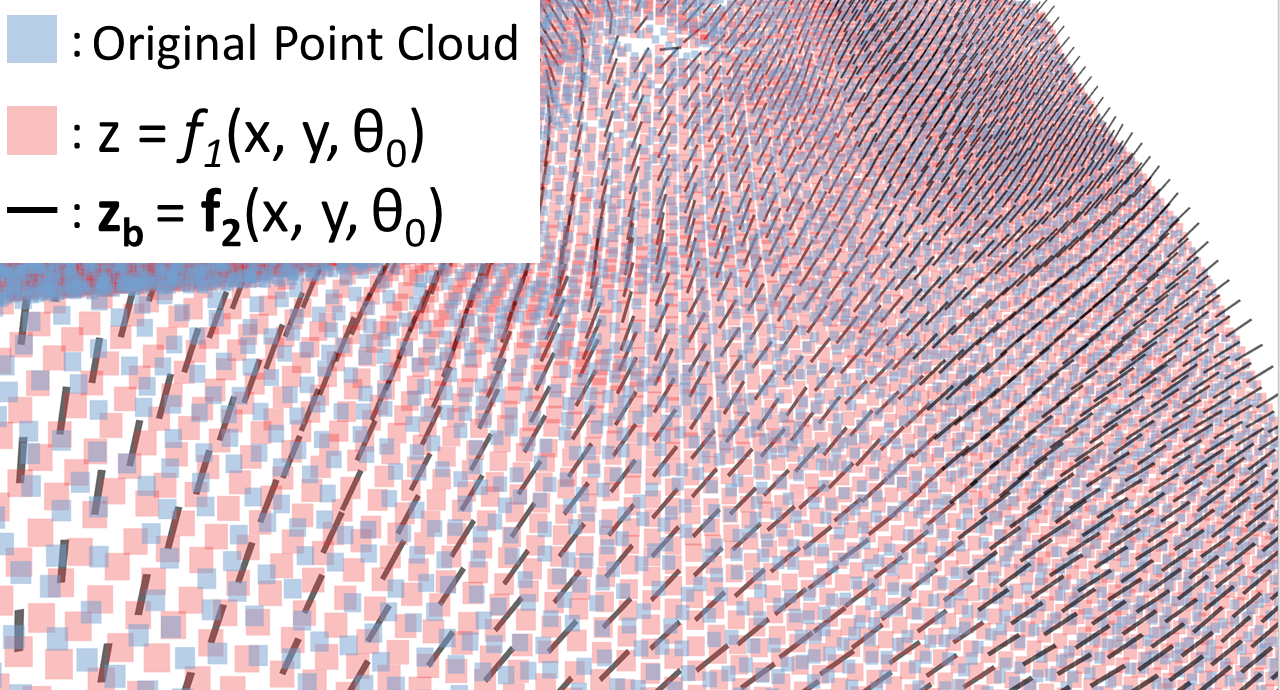}}
    % \quad
    \subcaptionbox{when $(x,y)=(x_c,y_c)$.}{\includegraphics[width=0.49\columnwidth]{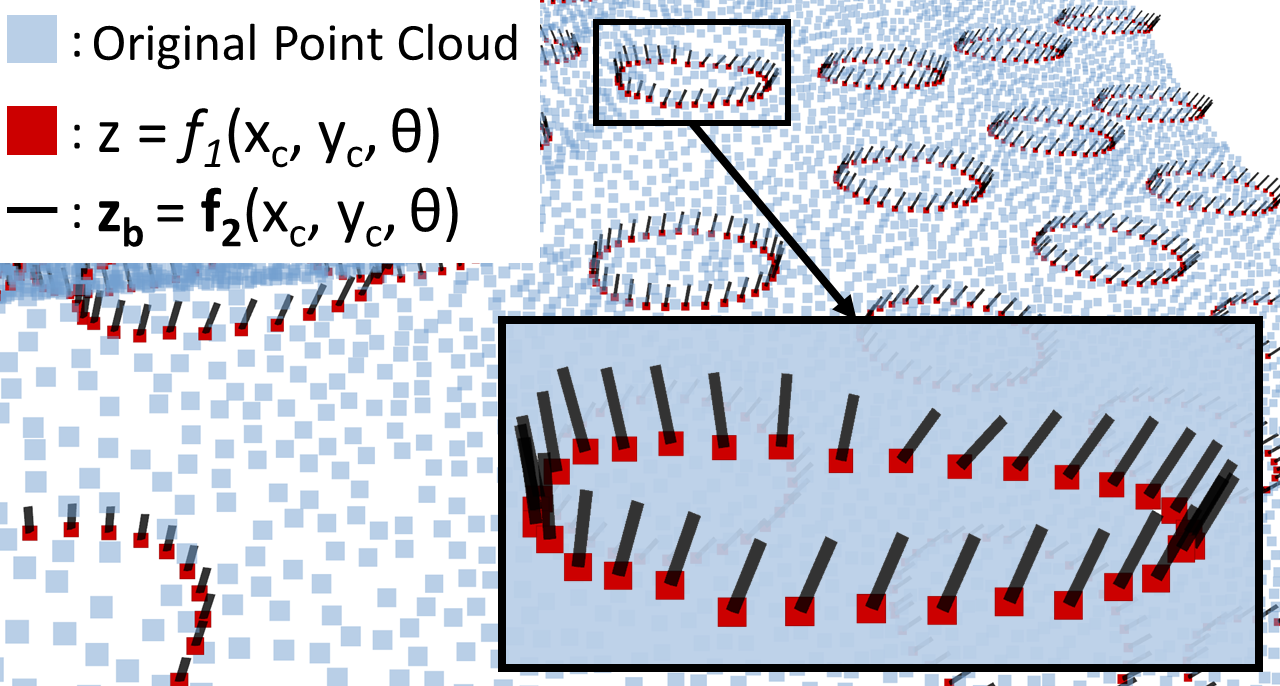}}
    \caption{The Process and Results of Constructing Mapping $\mathscr{F}$. Figure (a) illustrates the process of Algorithm \ref{alg:mapping construct}, where the gray vehicles indicate the robot poses obtained in each iteration. Figures (b) and (c) show the different $z$ and $\textbf{z}_b$ when $\theta$ or position $(x,y)$ is fixed, respectively, where the blue points are the original point cloud. The heights of the red points in Figure (b) and Figure (c) indicate the height $z\in\mathbb{R}$. The black line on each point indicates the $\textbf{z}_b\in\mathbb{S}^2_+$ with the direction pointing from the point to the sky. In Figure (c), black lines and red dot heights on each circle indicate $\textbf{z}_b$ and $z$ respectively when the position is the center of the circle $(x_c, y_c)$ but $\theta\in SO(2)$ is different. 
    Here, $\textbf{z}_b$ may vary with $\theta$ at the same position $(x_c, y_c)$, since the neighboring area of the robot may be rugged rather than a flat plane.}
    \vspace{-0.5cm}
    \label{fig:mapconstruct}
\end{figure}

We discretize $SE(2)$ space into grids and fit $\mathscr{F}$ for the state corresponding to each grid. When the value or gradient corresponding to a $SE(2)$ state is required, we then compute it using trilinear interpolation, where operations on manifold\cite{manifoldctrl} are used for the processing of $SO(2)$ and $\mathbb{S}_+^2$. Fig. \ref{fig:mapconstruct} illustrates the process and results of constructing mapping $\mathscr{F}$ through the example environments.

\subsection{Trajectory Parameterization}
\label{subsec:Trajectory Parameterization}
In this paper, we use quintic piecewise polynomials to represent state trajectories. Since the mapping $\mathscr{F}$ allows state space of the robot still be $SE(2)$, each piece of trajectory can be denoted as:
\begin{align}
&x_i(t)=c_{x_i}^\text{T}\gamma(t),\quad t\in[0,T_i];\\
&y_j(t)=c_{y_j}^\text{T}\gamma(t),\quad t\in[0,T_j];\\
&\theta_k(t)=c_{\theta_k}^\text{T}\gamma(t),\quad t\in[0,T_k];
\end{align}
where $i=1,2,...,N_i;j=1,2,...,N_j;k=1,2,...,N_k$ is the index of piecewise polynomial, $T_*,*=\{i,j,k\}$ is the duration of a piece of the trajectory, $c_*\in\mathbb{R}^6,*=\{x_i,y_j,\theta_k\}$ is the coefficient of polynomial, $\gamma(t)=[1,t,t^2,...,t^5]^\text{T}$ is the natural base.

We also add the constraint that the trajectory is four times continuously differentiable at the segmented points to obtain more continuous trajectories. Thus, let the whole trajectory be $\textbf{x}(t)=[x(t),y(t),\theta(t)]^\text{T}$, $\textbf{y}_{yaw}=[-\sin\theta,\cos\theta,0]^\text{T}$, combining with Eq.(\ref{model:pdot}) and Eq.(\ref{model:rdot}), we can compute the control inputs and dynamical variables analytically as follows:
\begin{align}
v_x&=\frac{v}{\cos\phi_x},\\
a_x&=\frac{a_t}{\cos\phi_x}+g\sin\varphi_x,\\
a_y&=\frac{a_n}{\cos\phi_y}+g\sin\varphi_y,\\
\omega_z&=\frac{\omega}{\cos\xi},\\
\kappa&=\frac{v_x}{\omega_z},\\
\delta&=\arctan(L_w\cdot\kappa),
\end{align}
where
\begin{align}
v&=\sqrt{\dot{x}^2+\dot{y}^2},\\
a_t&=\ddot{x}\cos\theta+\ddot{y}\sin\theta,\\
a_n&=-\ddot{x}\sin\theta+\ddot{y}\cos\theta,\\
\omega&=\dot\theta,\\
\cos\phi_{x}&=\textbf{x}_b^\text{T}\textbf{x}_{yaw},\quad \cos\phi_{y}=\textbf{y}_b^\text{T}\textbf{y}_{yaw},\\
\sin\varphi_x&=\textbf{x}_b^\text{T}\textbf{b}_{3},\quad\quad\sin\varphi_y=\textbf{y}_b^\text{T}\textbf{b}_{3},\\
\cos\xi&=\textbf{z}_b^\text{T}\textbf{b}_{3},
\end{align}
$a_x,a_y,\omega_z,\kappa$ denote longitude acceleration, latitude acceleration, angular velocity, and curvature, respectively. It is worth noting that $\cos\phi_x,\cos\phi_y,\sin\varphi_x,\sin\varphi_y,\cos\xi$ are all related to Eq.(\ref{con:zb}). When $\textbf{z}_b=\textbf{b}_3$, both control inputs and dynamical variables degenerate to the case that the ground is flat.

\subsection{Trajectory Optimization}
\label{subsec:Trajectory Optimization}
In this paper, we propose the cost function $\tau=\textbf{j}(t)^\text{T}\textbf{j}(t)+\rho_{\text{ter}}\cdot\sigma(\textbf{x}(t))$, where $\textbf{j}(t)=\textbf{x}^{(3)}(t)$ denotes the jerk of the trajectory, and its square integral represents the smoothness of the trajectory, $\rho_{\text{ter}}$ is a constant. $\sigma(\textbf{x}(t))$ is \textit{Surface Variation} proposed by \cite{surfcur} to approximate the terrain curvature, which can be obtained while calculating $\mathscr{F}$, since $\sigma(\textbf{x})=\lambda_0/\sum_{i=0}^{2}\lambda_i$, where $\lambda_0\leq\lambda_1\leq\lambda_2$ are eigenvalues of the covariance matrix of points in the ellipsoidal region near $\textbf{x}$.

We formulate the trajectory optimization problem for car-like robots on uneven terrain as:
\begin{align}
&\min_{\textbf{c}_{xy},\textbf{c}_{\theta},\textbf{T}_{xy},\textbf{T}_\theta}\int_0^{T_s}(\textbf{j}(t)^\text{T}\textbf{j}(t)+\rho_{\text{ter}}\sigma(\textbf{x}(t)))dt+\rho_TT_s\label{problem:opt}\\
&s.t.\ \ \dot x\sin\theta-\dot y\cos\theta=0,\label{con:non-holo}\\
&\quad\quad \textbf{M}_{xy}\textbf{c}_{xy}=\textbf{b}_{xy},\quad \textbf{M}_{\theta}\textbf{c}_{\theta}=\textbf{b}_{\theta},\label{con:mceqb}\\
&\quad\quad \textbf{T}_{xy}\succeq\textbf{0},\quad \quad\quad\ \ \textbf{T}_{\theta}\succeq\textbf{0},\label{con:tgeq0} \\
&\quad\quad v_x^2-v_{max}^2\leq0,\label{con:dyns}\\
&\quad \quad a_x^2-a_{mlon}^2\leq0,\\
&\quad \quad a_y^2-a_{mlat}^2\leq0,\\
&\quad \quad\frac{\omega_z^2}{v_x^2+\delta_+}-\frac{\tan^2\delta_{max}}{L_w^2}\leq 0,\label{con:dyne}\\
&\quad\quad c_{min}-\cos\xi\leq0,\label{con:attlim}\\
&\quad\quad\sigma(\textbf{x})-\sigma_{max}\leq0,\label{con:sigmalim}
\end{align}
where $\textbf{c}_{xy}=[[c_{x_1},c_{y_1}]^\text{T},[c_{x_2},c_{y_2}]^\text{T},...,[c_{x_M},c_{y_M}]^\text{T}]^\text{T}\in\mathbb{R}^{6M\times2},\textbf{c}_{\theta}=[c_{\theta_1}^\text{T},c_{\theta_2}^\text{T},...,c_{\theta_\Omega}^\text{T}]^\text{T}\in\mathbb{R}^{6\Omega\times1}$ are coefficient matrix, $M$ and $\Omega$ represent the number of pieces of the piecewise polynomial of $x,y$ and $\theta$, respectively. In this paper, we make the trajectory of theta have more pieces (i.e. $\Omega>M$) to better fit the non-holonomic constraint (\ref{con:non-holo}). 

$\textbf{T}_{xy}=[T_{1xy},T_{2xy},...,T_{Mxy}]^\text{T}\in\mathbb{R}^M$, $\textbf{T}_{\theta}=[T_{1\theta},T_{2\theta},...,T_{\Omega\theta}]^\text{T}\in\mathbb{R}^\Omega$ are time vectors, satisfying $\lVert\textbf{T}_{xy}\rVert_1=\lVert\textbf{T}_{\theta}\rVert_1=T_s$. Eq.(\ref{con:mceqb}) is combination of the continuity constraint mentioned in last subsection and boundary condition of the trajectory: 
\begin{align}
[\textbf{x}(0),\textbf{x}^{(1)}(0),\textbf{x}^{(2)}(0)]&=[\textbf{x}_{init},\textbf{x}^{(1)}_{init},\textbf{x}^{(2)}_{init}],\\
[\textbf{x}(T_s),\textbf{x}^{(1)}(T_s),\textbf{x}^{(2)}(T_s)]&=[\textbf{x}_{fina},\textbf{x}^{(1)}_{fina},\textbf{x}^{(2)}_{fina}], 
\end{align}
so $\textbf{M}_{xy}\in\mathbb{R}^{(5M+1)\times6M}$, $\textbf{M}_{\theta}\in\mathbb{R}^{(5\Omega+1)\times6\Omega}$.

Conditions (\ref{con:dyns})$\sim$(\ref{con:dyne}) are dynamic feasibility constrains, including limitation of longitudinal velocity $v_x$, longitude acceleration $a_x$, latitude acceleration $a_y$, and curvature $\kappa$, where $v_{max}^2,a_{mlon}^2,a_{mlat}^2,\delta_{max}$ are constants. $\delta_+$ in condition (\ref{con:dyne}) is a very small positive constant to avoid the zero denominators. Conditions (\ref{con:attlim}) and (\ref{con:sigmalim}) are limitations of attitude and terrain curvature, where $c_{min},\sigma_{max}$ are constants. We consider it unsafe for the robot to have too large $\xi$ or too large terrain curvature.

To deal with Eq.(\ref{con:mceqb}), we use the method proposed by work\cite{minco}, which allows eliminating Eq.(\ref{con:mceqb}) and converting the optimization variable $\textbf{c}_{xy},\textbf{c}_\theta$ to the segmentation point positions $\textbf{q}_{xy}\in\mathbb{R}^{(M-1)\times2}$ and $\textbf{q}_{\theta}\in\mathbb{R}^{\Omega-1}$, thus reducing the dimensionality of the problem. Moreover, we refer to the differential homogeneous mapping\cite{diffmorphi}, using the $C^2$ function mentioned in work\cite{diffcar} to map each element $T_{*}\in\mathbb{R}_+$ in $\textbf{T}_{xy}$ and $\textbf{T}_{\theta}$ to $\mathbb{R}$, eliminating the constraints (\ref{con:tgeq0}). As for the remaining non-holonomic constraint (\ref{con:non-holo}) and other inequality constraints (\ref{con:dyns})$\sim$(\ref{con:sigmalim}), we discretize each piece of the duration $T_{ixy}$ as $K$ time stamps $\tilde{t}_{ij}=(j/K)\cdot T_{ixy}$, $(i=1,2,...,M_{xy},j=0,1,...,K-1)$, and impose the remaining constraints on these time stamps, thus the number of constraints of the problem (\ref{problem:opt}) becomes $7KM_{xy}$. We also use this method to discretize $\sigma(\textbf{x}(t))$ and obtain its integral accumulatively.

Then, we use PHR Augmented Lagrange Multiplier method\cite{alm} (PHR-ALM) to solve the simplified problem, which is a method for solving constrained optimization problems, smoothing the dual function by adding quadratic terms, iteratively solving the approximate unconstrained problem and updating the dual variables. Besides, L-BFGS\cite{lbfgs} is chosen as the unconstrained optimization algorithm to work with PHR-ALM. It is a quasi-Newton method that can estimate the Hessian matrix from the previous objective function values and gradients with limited memory.

Last but not least, the gradients of the objective function and the constraints need to be calculated explicitly. It is easy to derive them using the chain rule, except for $\sigma$ and $\cos\phi_x,\cos\phi_y,\sin\varphi_x,\sin\varphi_y,\cos\xi$ with respect to $\textbf{f}_2$. For $\sigma$, since we can obtain it together with $\textbf{f}_2$, the gradient can be obtained using trilinear interpolation as well; for the others, let $\textbf{z}_b\triangleq[a,b,c]^\text{T}$, $\sin\theta\triangleq s_\theta$, $\cos\theta\triangleq c_\theta$, $r\triangleq c_\theta a+s_\theta b$, $s\triangleq as_\theta-bc_\theta$, we give their gradients with respect to state $\textbf{x}$ as follows:
\begin{align}
\nabla_\textbf{x}\cos\phi_x&=\nabla_\textbf{x}(\sqrt{1-r^2})=-r(1-r^2)^{-\frac{1}{2}}\nabla_\textbf{x}r,\\
\nabla_\textbf{x}\cos\phi_y&=\nabla_\textbf{x}(\frac{c}{\sqrt{1-r^2}})\\
&=(1-r^2)^{-\frac{1}{2}}\nabla_\textbf{x}c+r(1-r^2)^{-\frac{3}{2}}c\nabla_\textbf{x}r,\\
\nabla_\textbf{x}\sin\varphi_x&=\nabla_\textbf{x}(-\frac{rc}{\sqrt{1-r^2}})\\
&=-r(1-r^2)^{-\frac{1}{2}}\nabla_\textbf{x}c-(1-r^2)^{-\frac{3}{2}}c\nabla_\textbf{x}r\\
\nabla_\textbf{x}\sin\varphi_y&=\nabla_\textbf{x}(\frac{s}{\sqrt{1-r^2}}),\\
&=(1-r^2)^{-\frac{1}{2}}\nabla_\textbf{x}s+r(1-r^2)^{-\frac{3}{2}}s\nabla_\textbf{x}r,\\
\nabla_\textbf{x}\cos\xi&=\nabla_\textbf{x}(\textbf{b}_3^\text{T}\textbf{z}_b)=\nabla_\textbf{x}c.
\end{align}
In this work, we use the points in a unit circle on the X-Y plane to represent $\textbf{z}_b\in\mathbb{S}_+^2$, i.e., $a^2+b^2<1,c=\sqrt{1-a^2-b^2}$. Thus, $\nabla_\textbf{x}s,\nabla_\textbf{x}r$ are easily obtained since $\nabla_\textbf{x}a,\nabla_\textbf{x}b$ are obtained by trilinear interpolation. For $\nabla_\textbf{x}c$, due to $c=\sqrt{1-a^2-b^2}$, $\nabla_\textbf{x}c=-(a\nabla_\textbf{x} a+b\nabla_\textbf{x} b)/c$.

\section{Results}
\label{sec:Results}
\subsection{Implementation details}
\label{subsec:Implementation details}
In order to validate the performance of our method in real-world applications, we deploy it on a car-like robot, as shown in Fig. \ref{fig:real-world}. All computations are performed by an onboard computer NVIDIA Jetson Nano. We utilize NOKOV Motion Capture System\footnote{\url{https://en.nokov.com/}} for localization. Furthermore, a MPC controller\cite{mpc} with position feedback is fitted to the robot for trajectory tracking. We adopt the lightweight hybridA* algorithm as the front end of the planner and use Dubins Curve\cite{dubins} to shoot the end state for earlier termination of the search process. Besides, the implementation of the L-BFGS utilizes an open-source library LBFGS-Lite\footnote{\url{https://github.com/ZJU-FAST-Lab/LBFGS-Lite}}. All simulations are run on a desktop with an Intel i7-12700 CPU.

\subsection{Real-World Experiments }
\label{subsec:Real-World Experiments}
\begin{figure}	
    \centering
    \includegraphics[width=1.0\columnwidth]{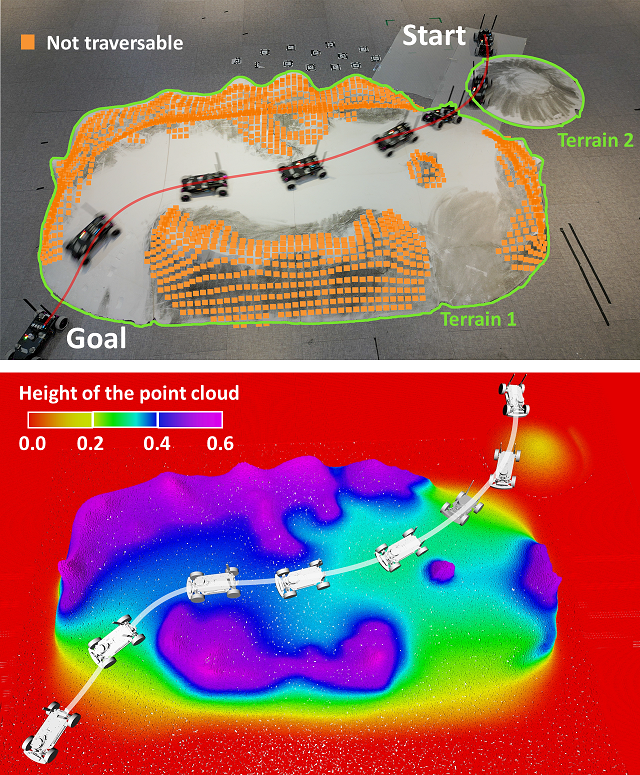}
	\caption{Real-World Experiments: In this case, the car-like robot needs to go downhill and uphill twice, traverse two terrains, and avoid the not traversable area determined by conditions (\ref{con:attlim}) and (\ref{con:sigmalim}). The length of the planned trajectory in 3D space (the red curve) is $5.496m$, planning time consuming is $1.603s$. The bottom half of this figure shows the visualization in RViz.}\label{fig:real-world}
    % \vspace{0.3cm}
\end{figure}

\begin{figure*}[t]  
		% \vspace{-0.0cm}  
		\centering
		{\includegraphics[width=1.0\linewidth]{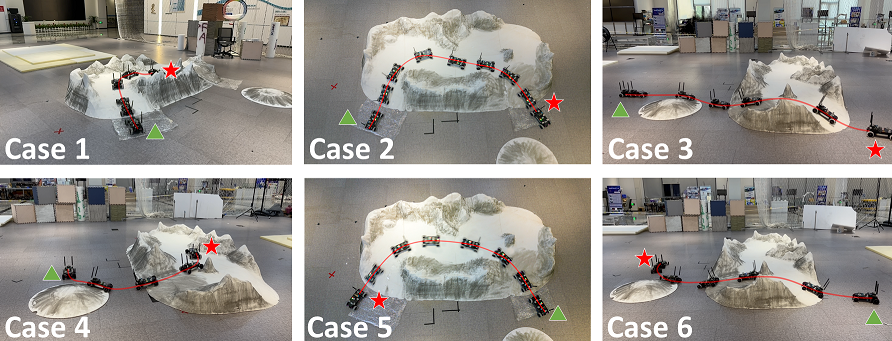}}
		\caption{The motion diagram of car-like robot in the real-world experiments, where the red curves are the planned trajectories. The green triangles indicate the starting points and the red pentagrams indicate the end points.}
		\label{fig:real_cases}
\end{figure*}

In the real-world experiments, as seen in Fig. \ref{fig:real-world}, we require the car-like robot to traverse two terrain models made of foam with distinct raised areas, obstacles, and slopes in order to test whether the proposed mapping $\mathscr{F}$ can help the planner generate safe and dynamically feasible trajectories for robots on uneven terrain. Limitation of longitudinal velocity, longitude acceleration, latitude acceleration, steering angle, attitude, and terrain curvature are set to $v_{max}=0.8m/s^2$, $a_{mlon}=5.0m/s^2$, $a_{mlat}=5.0m/s^2$, $\delta_{max}=0.505$, $c_{min}=0.86$ and $\sigma_{max}=0.05$, respectively. Meanwhile, we set the time weight $\rho_T=500$ and the terrain curvature weight $\rho_{\text{ter}}=10$ to ensure the aggressiveness of the trajectory. Fig. \ref{fig:real_cases} shows some cases with different starting and ending points in our tests.

Furthermore, to know the variation of relevant variables (e.g., attitude $\hat\xi$, forward velocity $\hat{v}_{x}$, etc.) during the motion of the robot, we make statistics in Tab. \ref{tab:real_dyn} for the case shown in Fig. \ref{fig:real-world}. As we can see, the robot maintains a relatively high speed and small tracking error throughout to reach the target state without violating the constraints we set. More demonstrations can be found in the attached multimedia.

\begin{figure}	
    \centering
    \includegraphics[width=1.0\columnwidth]{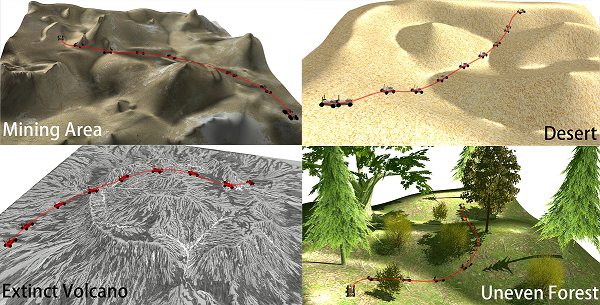}
	\caption{A car-like robot navigating autonomously through different uneven terrain.}\label{fig:sim-exp}
    % \vspace{0.3cm}
\end{figure}

\subsection{Simulation Experiments }
\label{subsec:Simulation Experiments}
To testify the effectiveness of our method in more extensive and complex environments, we build an open-source simulation environment based on Gazebo\footnote{\url{https://gazebosim.org/}}. There are some typical uneven terrain, including deserts, mining area, volcanic area, etc. We simulate a car-like robot driving in different environments. For example, in the uneven forest, the robot needs to avoid obvious obstacles such as shrubs and trees; in an extinct volcanic environment, the robot needs to find a flatter entrance when crossing the crater. As for desert and mining area, the robot must maintain a balance between traversability and path length while planning the trajectory to traverse them. Fig. \ref{fig:sim-exp} shows the robot navigating autonomously through different simulation scenarios.  

\begin{table}[t]
	\small
	\centering
	\renewcommand\arraystretch{1.15}
	\caption{\label{tab:real_dyn} Statistics in Real-World Experiments }
	\begin{tabular}{lclll}
		\toprule
		Statistics & Mean  &  Max & STD.\\
		\midrule
		$\hat{v}_{x}$ ($m/s$) &0.684  &0.800 &0.212	\\
		\midrule
		$\hat\xi$ ($rad$) &0.117  &0.366  &0.108 \\
		\midrule
		  $\text{Tracking Error}$ ($m$) &0.059  &0.109  &0.025 \\
		\bottomrule
	\end{tabular}
\end{table}

\begin{figure}
    \centering
    \subcaptionbox{Trajectory visualization in mountainous terrain. \label{fig:traj_visual}}{
        \includegraphics[width=1.0\columnwidth]{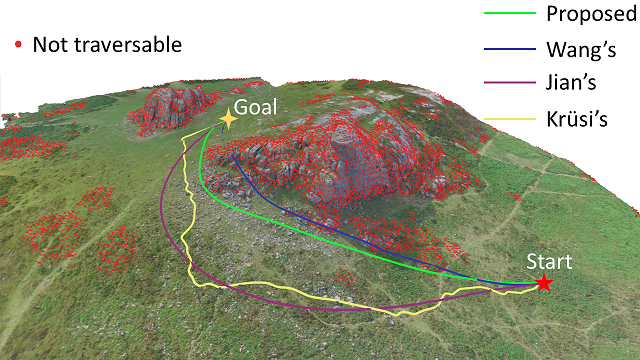}}
	\newline
	\vspace{0.3cm}
    \subcaptionbox{Comparison of velocity curves of trajectories.\label{fig:traj_vel}}{
        \includegraphics[width=1.0\columnwidth]{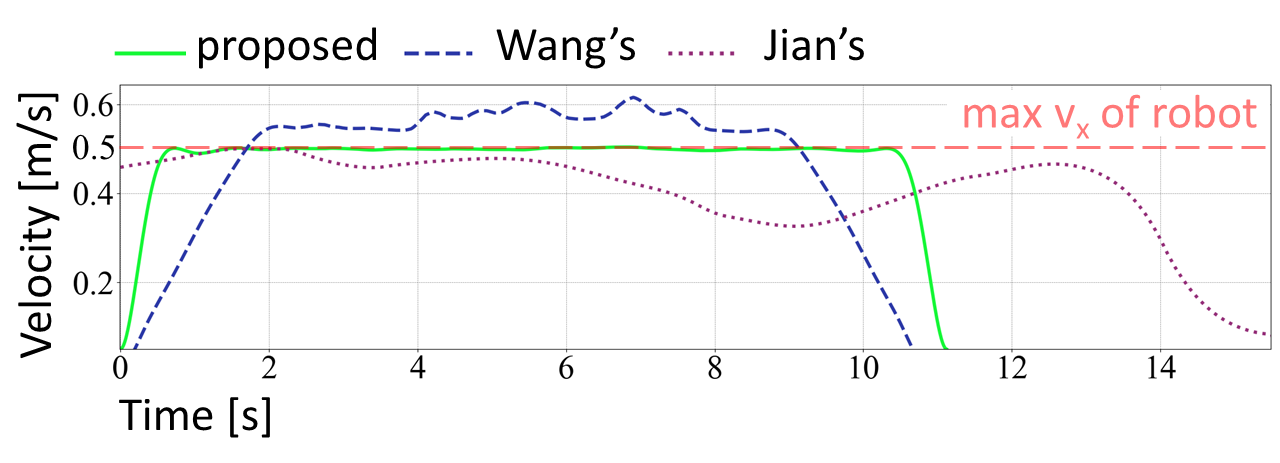}}
    \caption{Trajectory Visualization and Comparison}
\end{figure}

\subsection{Benchmark Comparisons}
\label{subsec:Benchmark Comparisons}
In this subsection, we compare the proposed planning algorithm with Kr{\"u}si's\cite{drivingonpoint}, Jian's\cite{putn} and Wang's\cite{towards} methods. We first compared the velocity curve of the trajectory with the last two works. The comparison is conducted on mountainous terrain, as shown in Fig. \ref{fig:traj_visual}. In addition, the maximum velocity and longitude acceleration were set to $v_{max}=0.5m/s$ and $a_{mlon}=5.0m/s^2$, respectively. As can be seen from the curves in Fig. \ref{fig:traj_vel}, our trajectory planner can better utilize the maneuverability of the robot. The trajectory planned by Wang's method\cite{towards} is shorter. However, because their method does not consider the terrain contact constraint, it is easy to optimize the trajectory to a not traversable area. Moreover, the generated trajectory violates the constraint of maximum velocity($v_x$) of the robot since the dynamics coupled with the changing terrain is not taken into account. Regarding Jian's method\cite{putn}, although the changing terrain is modeled in the NMPC problem, for two reasons, the trajectory planned by their method is less optimal and has a longer execution time and length. The first reason is that they do not consider the execution time in the optimization. The second one is the insufficient iterations of the sampling-based front end.

To further quantitatively measure the performance of the trajectory with existing methods, some generic evaluation metrics are used to compare, including the mean of planning time consuming ($t_{p}$), mean curvature of the trajectory ($\kappa_m$) which reflects the smoothness of the trajectory, and mean of tracking error ($\text{Tra}_{err}$). We chose four scenes in our simulation environment mentioned in Sec. \ref{subsec:Simulation Experiments}. All parameters are finely tuned for the best performance of each method.

\begin{table}[t]
    \vspace{-0.4cm}
	\small
	\centering
	\renewcommand\arraystretch{1.2}
	\caption{\label{tab:bk} Benchmark Comparison }
	\begin{tabular}{c|clllllllll}
		\hline
		\multirow{2}{*}{Method}& \multirow{1}{*}{$t_{p}(s)$} & \multicolumn{1}{c}{$\kappa_m(m^{-1})$} &  \multicolumn{1}{c}{$\text{Tra}_{err}(m)$}\\
		& \multicolumn{1}{c}{ } & \multicolumn{1}{c}{ } & \multicolumn{1}{c}{ }\\ \hline
        proposed
		& \textbf{0.301} &\textbf{0.710} &\textbf{0.086}\\
		Kr{\"u}si's\cite{drivingonpoint}                 
		&4.696 &1.530 & - \\
            Jian's\cite{putn}                 
		&0.450 &0.910 &0.101\\
		Wang's\cite{towards}                 
		&0.960 &0.717 &0.117\\ \hline
	\end{tabular}
	\vspace{-0.2cm}
\end{table}

More than one thousand comparison tests are performed in each scene with random starting and ending states. Since the trajectory generated by Kr{\"u}si's method\cite{drivingonpoint} does not provide time-related information, we did not count its $\text{Tra}_{err}$. The result is summarized in Table \ref{tab:bk}. It states that mean planning time of Kr{\"u}si's method\cite{drivingonpoint} is much longer than that of other three. This is because the maximum curvature constraint of the robot limits the size of the trajectory library used in this method. Wang's method\cite{towards} also has a long planning time due to the absence of a better heuristic function for graph search. Our trajectories are smooth and have a small mean curvature since the polynomial is used to represent the trajectory and the norm of its high-order derivatives is minimized. Moreover, thanks to our rational formulation of the problem and optimization strategy providing better continuity of the robot states and their finite-dimensional derivatives, as well as our effective consideration of traversability using terrain curvature, our trajectories are easier to track by the controller.

In Table \ref{tab:bk}, our method achieves better performance in terms of mean of planning time consuming $t_{p}$, mean curvature of the trajectory $\kappa_m$ and mean of tracking error $\text{Tra}_{err}$, which shows that the algorithm we propose is more efficient and effective. The trajectories generated based on the proposed planning framework have higher quality.

\section{Conclusion}
\label{sec:Conclusion}
In this paper, we propose the mapping $\mathscr{F}:SE(2)\mapsto\mathbb{R}\times\mathbb{S}^2_+$ to describe the impact of terrain on the robot and present an efficient algorithm to construct it approximately. Using this mapping to model the motion of a car-like robot on uneven terrain, we present a trajectory optimization framework for car-like robots on uneven terrain that can consider terrain curvature and dynamics of the robot. Real-world experiments and simulation benchmark comparisons validate the efficiency and quality of our method. In fact, more factors can also be considered in this framework in the form of constraints, such as obstacles and user-defined risks.

However, there is still space for improvement in our method. We did not consider the suspension system of the robot, the interaction of the robot's tires with the ground and the possible tire skidding. Also, in more complex scenarios (e.g., weedy land, deserts, muddy land), it is challenging to construct accurate dynamics models of the robot in contact with the terrain, which may require a combination of other tools, such as neural networks, stochastic processes, etc.

In the future,  we will extend this algorithm to multi-layer environments and more difficult field environments. Besides, to further exploit the advantages of the efficiency of our method, local re-planning will be considered and adapted to the dynamic environments.

\newlength{\bibitemsep}\setlength{\bibitemsep}{0.00\baselineskip}
\newlength{\bibparskip}\setlength{\bibparskip}{0pt}
\let\oldthebibliography\thebibliography
\renewcommand\thebibliography[1]{
    \oldthebibliography{#1}
    \setlength{\parskip}{\bibitemsep}
    \setlength{\itemsep}{\bibparskip}
}
\bibliography{references}

\end{document}